\documentclass{article} 
\usepackage[preprint]{neurips_2026}

\usepackage{amsmath,amsfonts,bm}

\def\eqref#1{equation~\ref{#1}}

\def\1{\bm{1}}

\DeclareMathAlphabet{\mathsfit}{\encodingdefault}{\sfdefault}{m}{sl}
\SetMathAlphabet{\mathsfit}{bold}{\encodingdefault}{\sfdefault}{bx}{n}

\usepackage[T1]{fontenc}
\usepackage{microtype}
\usepackage{url}
\usepackage{graphicx}
\graphicspath{{figures/}}
\usepackage{booktabs}
\usepackage{amsmath}
\usepackage{amssymb}
\usepackage{array}
\usepackage{placeins}
\usepackage{float}
\usepackage{hyperref}
\hypersetup{hidelinks}
\newcolumntype{L}[1]{>{\raggedright\arraybackslash}p{#1}}
\usepackage{xcolor}
\usepackage{amsthm}
\usepackage{enumitem}

\title{Function-Space Transformer with Adaptive Anchors}

\author{Guorui Sang \\ University of Illinois Chicago\\gsang@uic.edu \And
Pedram Rooshenas \\ University of Illinois Chicago\\pedram@uic.edu}

\begin{document}

\maketitle

\begin{abstract}
Many forms of data, including physical fields, geometric shapes, and visual signals, are naturally described by functions over continuous domains but are observed through discrete samples. Representing these functions on fixed uniform grids imposes a trade-off between resolving localized variation and increasing computation across the domain. Neural operators address this mismatch by learning mappings between functions, while latent-attention architectures provide flexible processing of sampled observations. We introduce the Function-Space Transformer (FST), a framework for learning from functions through a spatially adaptive continuous latent representation. FST stores features at anchors whose locations are predicted from the input observations and recursively refines these anchor features through function-space interactions. This allows the representation to adapt its spatial organization to each input rather than inherit that of the observation grid, while supporting both spatially resolved and finite-dimensional outputs. On PDE solution prediction using PDEBench Burgers and Darcy flow, FST substantially outperforms the Perceiver IO baseline, whose latent representation lacks explicit spatial organization, and is highly competitive with the Fourier Neural Operator. On ImageNet-1K, FST achieves higher classification accuracy than the Vision Transformer baseline, with fewer parameters across these comparisons. Ablations further support the benefits of function-space updates and recursive refinement. Together, these results highlight the potential of adaptive continuous representations for both scientific prediction and visual recognition.
\end{abstract}

\section{Introduction}
Applying transformers to functions over continuous domains requires representing the input through a finite collection of tokens. In common grid- and patch-based formulations, the spatial organization of these tokens follows the input discretization~\citep{ViT}. This couples representational resolution to computational cost: finer tokenizations can preserve smaller-scale variation, but increasing the number of tokens raises the quadratic cost of dense self-attention~\citep{Perceiver}. Coarser tokenizations reduce this cost but may obscure localized structure needed for prediction. This trade-off is especially restrictive when spatial variation is nonuniform. Resolving a narrow transition or fine boundary through uniform refinement increases computation throughout the domain, including regions that require little additional detail.

The discretization used to observe a function, however, need not determine how a model represents it internally. Even when the input is sampled on a fixed grid, latent features need not follow the same layout or allocate equal capacity to equally sized regions. Instead, their spatial organization can adapt to the structure of each input. The challenge is therefore not simply to process a fixed set of tokens more efficiently, but to learn how to distribute a finite representation across a continuous domain. This motivates a continuous latent field whose spatial allocation is learned from the observations rather than prescribed by their sampling grid.

Neural operators approach learning from discretely observed functions by modeling mappings between function spaces. The Fourier Neural Operator (FNO;~\citealp{FNO}), for example, models global interactions through spectral transformations, while attention can also be formulated over continuous domains~\citep{ContinuumAttention}. Latent-attention architectures such as Perceiver~\citep{Perceiver} and Perceiver IO~\citep{PerceiverIO} offer a complementary approach: they encode sampled observations into a smaller latent array, reducing the cost of processing large inputs and supporting task-dependent outputs. However, their latent arrays lack an explicit spatial organization that defines a continuous field. Spatial representations such as Equivariant Neural Fields (ENFs;~\citealp{ENF}) associate latent features with locations and use Gaussian windows to define a field between them. These perspectives motivate combining compact latent processing with a continuous representation whose spatial organization adapts to each input.

We introduce the \emph{Function-Space Transformer} (FST), which represents each input through a continuous latent field defined by input-adaptive \emph{anchors}. Each anchor pairs a coordinate predicted from the observations with a latent feature, and a weighted blend of the anchor features defines the field throughout the domain. This construction allows the model to learn where to store latent features rather than inherit their placement from the observation grid. The resulting representation supports both spatially resolved outputs, such as PDE solution fields, and finite-dimensional outputs, such as class predictions.

FST refines this representation through recursive function-space interactions. At each refinement step, the anchors first attend to the input observations. A function-space layer then evaluates the latent field at sampled points, applies global self-attention among the evaluated features, and writes the processed information back to the anchors through cross-attention. This separates the coordinates used to store the representation from those used to compute its global interactions. Rather than applying global attention directly to stored spatial latents~\citep{ENFPDE,UPT,ABUPT} or modal tokens~\citep{GaussianParticleOperator}, FST processes sampled evaluations of the continuous field and uses them to update its stored features. Parameters are shared across refinement steps, similar to the weight-sharing approaches used by \citet{UniversalTransformer} and \citet{TRM}. The input-predicted anchor coordinates remain fixed during refinement, while the anchor features and the field they define evolve. Our contributions are:
\begin{itemize}[leftmargin=*,itemsep=1pt,topsep=2pt]
\item \textbf{Adaptive continuous representation.}
We introduce FST, which constructs a continuous latent field from features stored at input-predicted anchors. This allows the spatial organization of the representation to adapt to each input rather than follow the observation grid, while supporting both spatially resolved and finite-dimensional outputs.

\item \textbf{Function-space interactions and recursive refinement.}
We propose a shared function-space layer that applies global self-attention to sampled evaluations of the latent field and transfers the resulting information back to its anchors. This separates representation storage from interaction sampling and recursively refines the field with repeated access to the observations.

\item \textbf{Experimental evaluation.}
FST substantially outperforms the Perceiver IO baseline on Burgers and Darcy solution prediction and achieves accuracy comparable to FNO on all-viscosity Burgers and Darcy~\citep{PDEBench,PerceiverIO,FNO}. On ImageNet-1K, it achieves higher classification accuracy than the ViT baseline~\citep{ViT}. FST uses fewer parameters across these comparisons. Ablations examine adaptive anchors and support the benefits of function-space updates and recursive refinement.
\end{itemize}

\section{Related Work}

\noindent\textbf{Attention-based learning in function space.}
Continuum attention defines attention as an operator between function spaces and studies its discrete approximation from sampled values~\citep{ContinuumAttention}. Perceiver aggregates observations into a fixed number of latent features and applies self-attention among them~\citep{Perceiver}; Perceiver IO adds output queries, separating latent computation from the output structure~\citep{PerceiverIO}. For operator learning, the Inducing
Point Operator Transformer and Universal Physics Transformers (UPT) use this
separation to process latent features and predict function values at specified
coordinates~\citep{IPOT,UPT}. FST inherits this separation of latent processing and output queries; its difference lies in how the latent representation is organized and where interactions are computed.

\noindent\textbf{Learning with spatially structured representations.}
LIIF, Convolutional Occupancy Networks, and GINO decode from or connect to latent features on fixed regular grids~\citep{LIIF,ConvONet,GINO}; deformable attention adapts sampling locations while retaining a regular output grid~\citep{DAT}. ENFs use Gaussian windows to localize the contributions of spatial latent features and fit these features and their geometry to each signal's observations~\citep{ENF}. Related continuous representations also appear in CViT~\citep{CViT} and AROMA~\citep{AROMA}. FST instead predicts input-adaptive anchor coordinates in a single forward pass and recursively refines their features for the target task.

These spatial representations also differ in how they exchange information (Table~\ref{tab:positioning}). For PDE forecasting, ENFs use message passing among pose--feature pairs~\citep{ENFPDE}. GPO predicts Gaussian geometry, pools the representation into modal tokens, applies attention among them, and scatters the result back to spatial locations~\citep{GaussianParticleOperator}. FST applies attention directly to features evaluated at sampled points, whose number can be chosen separately from the anchor count. AB-UPT applies self-attention among anchor tokens, while query tokens read from the anchors to produce independent predictions~\citep{ABUPT}. In FST, self-attention mixes the evaluated features, and cross-attention writes the result back to the anchor features before output decoding.

\begin{table}[t]
\caption{Where each architecture stores its latent representation and where it computes global interaction.}
\label{tab:positioning}
\centering\small
\setlength{\tabcolsep}{4pt}
\begin{tabular}{@{}L{0.17\linewidth}L{0.33\linewidth}L{0.44\linewidth}@{}}
\toprule
Model & Latent representation & Global interaction \\
\midrule
Perceiver IO / UPT & Unstructured latent tokens & Self-attention among latent tokens \\
ENF & Latent features optimized for each input with Gaussian windows & Message passing among latents (ENF-PDE) \\
GPO & Predicted Gaussian geometry & Attention among modal tokens \\
AB-UPT & Spatial anchor tokens; separate query tokens & Self-attention among anchors; queries only read anchors \\
FST (ours) & Anchors predicted from the input, normalized RBFs & Attention among sampled field evaluations, written back \\
\bottomrule
\end{tabular}
\end{table}

\noindent\textbf{Recursive refinement.}
Perceiver repeatedly attends to the observations because a fixed latent can omit needed detail~\citep{Perceiver}. Universal Transformers and recent recursive models refine latent states with a shared network~\citep{UniversalTransformer,HRM,TRM}, increasing depth through parameter sharing. FST uses this recursive approach to refine the anchor features.

\vspace{-0.09in}
\section{Method}
\label{sec:method}
\vspace{-0.09in}

Let \(\Omega\subseteq\mathbb{R}^{d_x}\) be a domain. We learn a map \(\mathcal{G}:\mathcal{F}(\Omega;\mathbb{R}^{d_f})\rightarrow\mathcal{Y}\), where \(\mathcal{Y}\) is a function space \(\mathcal{F}(\Omega;\mathbb{R}^{d_g})\) or \(\mathbb{R}^{C}\). The model accesses the input \(f\) through \(N\) observations: \(\mathcal{O}=\{(\mathbf{x}_i,\mathbf{v}_i)\}_{i=1}^{N}\), with values \(\mathbf{v}_i\in\mathbb{R}^{d_v}\) at \(\mathbf{x}_i\in\Omega\). For a function output \(g=\mathcal{G}(f)\), the model predicts \(\widehat{g}(\mathbf{q}_k)=\mathcal{G}_{\theta}(\mathcal{O})(\mathbf{q}_k)\) at output queries \(\mathcal{Q}=\{\mathbf{q}_k\}_{k=1}^{Q}\); otherwise it returns \(\widehat{\mathbf{y}}=\mathcal{G}_{\theta}(\mathcal{O})\in\mathbb{R}^{C}\).

\begin{figure}[t]
  \centering
  \includegraphics[width=0.95\linewidth]{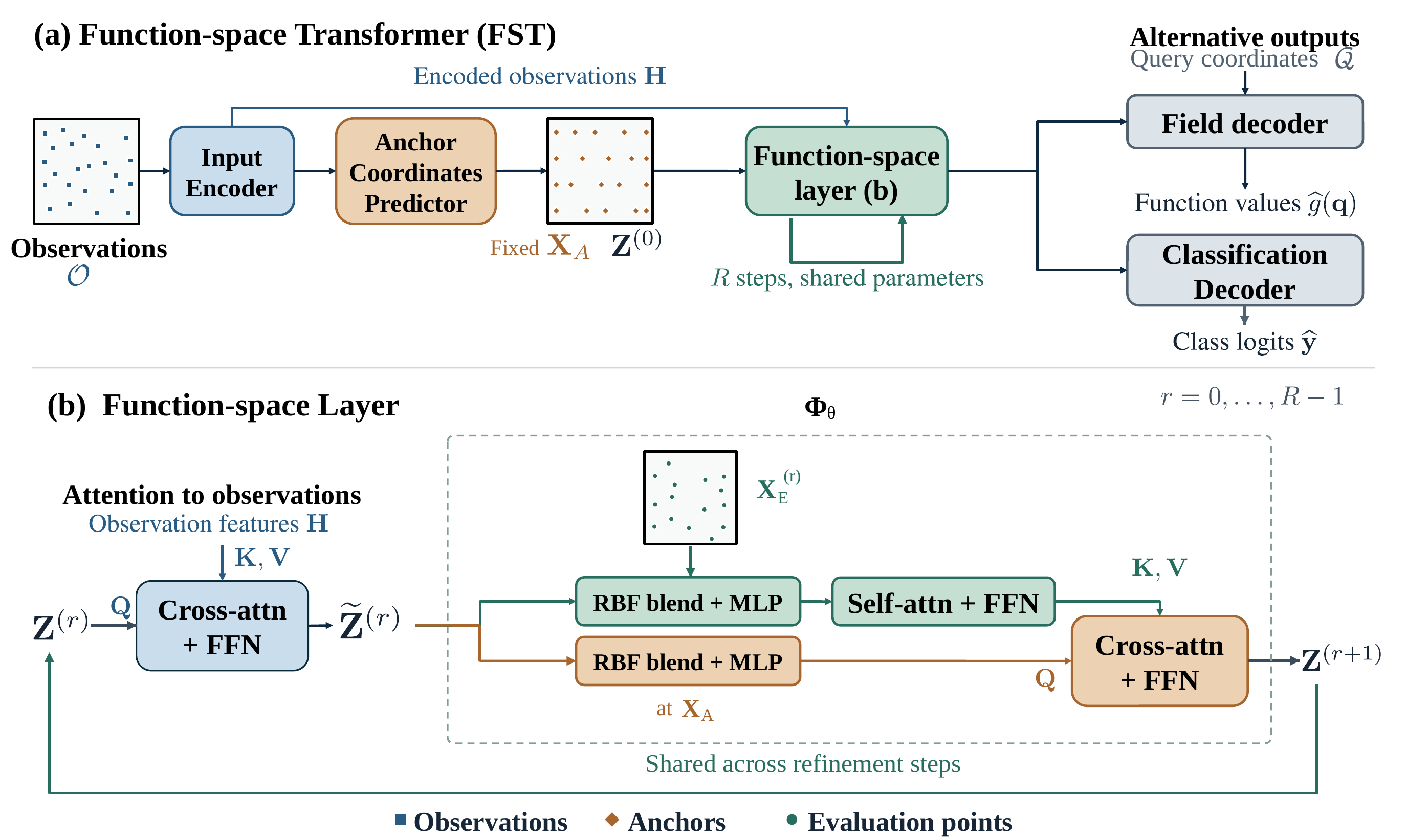}
  \caption{Function-Space Transformer (FST). (a) A shared layer refines anchor features for \(R\) steps at fixed coordinates \(\mathbf X_A\), predicted once from the input. The decoder queries the field at \(\mathcal Q\) or pools anchor features. (b) Each step attends to observations, evaluates RBF blends through a pointwise MLP at \(\mathbf X_E^{(r)}\), and applies self-attention. Cross-attention updates anchors using queries from the same RBF blend and MLP at \(\mathbf X_A\).}
  \label{fig:method-overview}
\end{figure}

FST constructs a continuous latent field from input-adaptive anchors (Figure~\ref{fig:method-overview}). A shared function-space layer refines this field \(R\) times. Each step attends to the observations, mixes features evaluated at sampled points, and writes the result back to the anchor features. The final anchor features are decoded at output queries for function prediction or pooled to produce a vector output.

FST uses three coordinate sets: \(M\) anchors at \(\mathbf X_A\) store features \(\mathbf Z\); \(S\) evaluation points at \(\mathbf X_E\) specify where self-attention operates; and \(Q\) output queries in \(\mathcal Q\) specify where function predictions are requested. These sets share a domain and may overlap.

\noindent\textbf{Anchor Representation.} Different inputs, such as PDE solutions with fronts at different locations, call for different spatial allocations. FST therefore predicts anchor coordinates from the observations. The coordinate predictor runs once per input and also supplies length scales when these are adaptive. Coordinates remain fixed during refinement (Appendix~\ref{app:coordinate-parameterization}).

Each anchor pairs a coordinate \(\mathbf{a}_j\in\Omega\) with an anchor feature \(\mathbf{z}_j\in\mathbb{R}^{d_z}\), giving the anchor set
\(\mathcal{A}=\{(\mathbf{a}_j,\mathbf{z}_j)\}_{j=1}^{M}\).
The anchor features are initialized from a coordinate embedding \(\gamma\) and an optional global condition \(\mathbf c\) as
\(\mathbf{z}_j^{(0)}=C_{\theta}(\gamma(\mathbf{a}_j),\mathbf c)\).

\noindent\textbf{Field construction.}
FST defines a continuous latent field using normalized Gaussian radial basis functions (RBFs), which construct continuous approximations from scattered coordinates~\citep{BuhmannRBF}:
\begin{equation}
w_j(\mathbf{x})
=
\frac{
\exp\!\left(-\tfrac12(\mathbf{x}-\mathbf{a}_j)^{\top}\Lambda_j^{-1}(\mathbf{x}-\mathbf{a}_j)\right)}
{\sum_{m=1}^{M}\exp\!\left(-\tfrac12(\mathbf{x}-\mathbf{a}_m)^{\top}\Lambda_m^{-1}(\mathbf{x}-\mathbf{a}_m)\right)},
\qquad
\bar{\mathbf{z}}(\mathbf{x})
=
\sum_{j=1}^{M}w_j(\mathbf{x})\mathbf{z}_j.
\label{eq:rbf}
\end{equation}
We call this RBF-weighted average of anchor features an \emph{RBF blend}. Here, \(\Lambda_j=\operatorname{diag}(\ell_{j,1}^2,\ldots,\ell_{j,d_x}^2)\) contains the RBF length scales of anchor \(j\). They set the spatial spread of each Gaussian~\citep[Section~3.2]{Cavoretto2021}; setting \(\Lambda_j=\ell_j^2\mathbf{I}\) gives isotropic weights. RBF length scales are fixed, predicted from the observations, or learned as input-independent model parameters. We denote the weights used during refinement by \(w_j^{\mathrm{ref}}\) and their length scales by \(\Lambda^{\mathrm{ref}}\).

\subsection{Function-Space Layer}
\label{sec:fs-layer}

The function-space layer updates the anchor features and hence the latent field while keeping anchor coordinates fixed.

Let \(\mathbf{X}_A\in\mathbb{R}^{M\times d_x}\) and \(\mathbf{Z}^{(r)}\in\mathbb{R}^{M\times d_z}\) denote the anchor coordinates and the anchor features at refinement step \(r\), and let \(\mathbf{H}\in\mathbb{R}^{N\times d_z}\) denote the encoded observations. Each refinement step has three stages.

\noindent\textbf{(i) Attention to observations.}
As in Perceiver~\citep{Perceiver}, anchors first attend to observations. An optional bias \(\mathbf B^{\mathrm{obs}}\in\mathbb R^{M\times N}\) depends on anchor and observation coordinates:
\begin{equation}
\begin{aligned}
\mathbf{U}^{(r)}
&=
\mathbf{Z}^{(r)}
+
\operatorname{MHA}_{\mathbf B^{\mathrm{obs}}}\!\left(
\operatorname{LN}_{A}(\mathbf{Z}^{(r)}),
\operatorname{LN}_{H}(\mathbf{H}),
\operatorname{LN}_{H}(\mathbf{H})
\right),\\
\widetilde{\mathbf{Z}}^{(r)}
&=
\mathbf{U}^{(r)}
+
\operatorname{FFN}\!\left(
\operatorname{LN}_{2}(\mathbf{U}^{(r)})
\right),
\end{aligned}
\label{eq:obs-read}
\end{equation}
with layer normalization (LN), a feed-forward network (FFN), and multi-head attention \(\operatorname{MHA}_{\mathbf B}\) whose logits are \(\mathbf{Q}\mathbf{K}^\top/\sqrt{d}+\mathbf B\), where \(d\) is the key dimension per head.

\noindent\textbf{(ii) Evaluate and mix.}
\label{sec:sampling}
Let \(\mathbf{X}_E^{(r)}=[\mathbf{e}_1;\ldots;\mathbf{e}_S]\in\mathbb{R}^{S\times d_x}\) contain the evaluation coordinates for step \(r\). Evaluation points can cover the domain uniformly or follow the anchor distribution to emphasize regions with denser anchors. We combine these choices for PDE prediction. The points may be fixed or resampled across different steps.

We blend the anchor features at these coordinates and apply a pointwise MLP:
\begin{equation}
\begin{aligned}
W^{(r)}_{sj}&=w^{\mathrm{ref}}_j(\mathbf{e}_s),\\
\mathbf{E}^{(r)}&=F_{\theta}\!\left(W^{(r)}\widetilde{\mathbf{Z}}^{(r)},\gamma(\mathbf{X}_E^{(r)}),\mathbf c\right)\in\mathbb{R}^{S\times d_z},
\end{aligned}
\label{eq:evaluation}
\end{equation}
where \(F_{\theta}\) maps each RBF blend and coordinate embedding to an evaluated feature, with an optional global condition \(\mathbf c\). Self-attention then mixes the \(S\) evaluated features:
\begin{equation}
\begin{aligned}
\mathbf{G}^{(r)}
&=
\mathbf{E}^{(r)}
+
\operatorname{MHA}\!\left(
\operatorname{LN}_{E}(\mathbf{E}^{(r)}),
\operatorname{LN}_{E}(\mathbf{E}^{(r)}),
\operatorname{LN}_{E}(\mathbf{E}^{(r)})
\right),\\
\widetilde{\mathbf{G}}^{(r)}
&=
\mathbf{G}^{(r)}
+
\operatorname{FFN}\!\left(\operatorname{LN}_{G}(\mathbf{G}^{(r)})\right).
\end{aligned}
\label{eq:mix}
\end{equation}

\begin{figure}[t]
  \centering
  \includegraphics[width=0.9\linewidth]{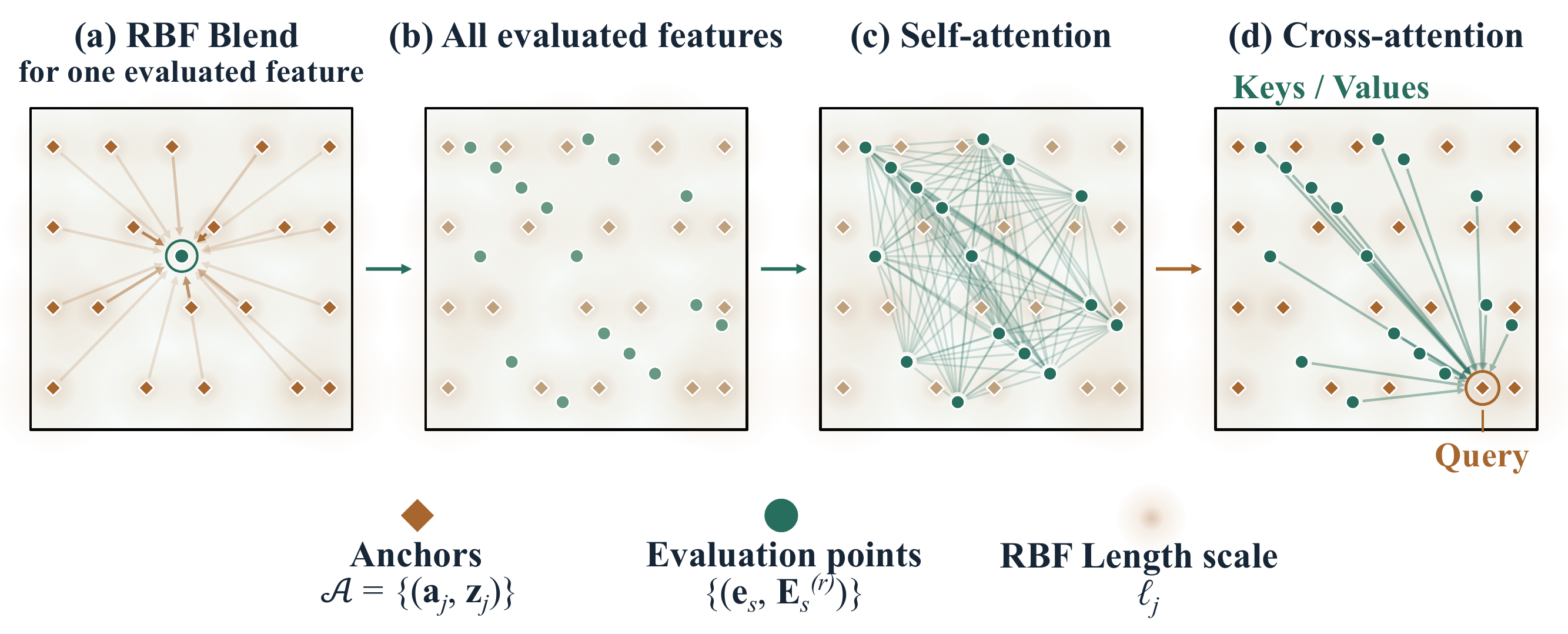}
  \caption{Evaluate, mix, and write-back. (a) RBF weights blend anchor features; a pointwise MLP, omitted for clarity, maps the blend and coordinate embedding to an evaluated feature. (b) Evaluation points may include anchor coordinates. (c) Self-attention mixes evaluated features. (d) Cross-attention incorporates information into anchor features. Its queries use the same RBF blend and MLP at anchor coordinates. Anchor coordinates stay fixed; one receiving anchor is highlighted.}
  \label{fig:evaluate-mix-write-back}
\end{figure}

\noindent\textbf{(iii) Write back.}
Cross-attention writes information from the mixed features back to the anchor features. To form the queries, we apply the same RBF blending and pointwise MLP at the anchor coordinates. Even at \(\mathbf a_j\), the blend combines features from all anchors, so \(\bar{\mathbf z}(\mathbf a_j)\neq\mathbf z_j\) in general. With \(W^{A}_{jm}=w^{\mathrm{ref}}_m(\mathbf{a}_j)\),
\begin{equation}
\begin{aligned}
\mathbf{Q}_A^{(r)}
&=F_{\theta}\!\left(W^{A}\widetilde{\mathbf{Z}}^{(r)},\gamma(\mathbf{X}_A),\mathbf c\right),\\
\mathbf{V}^{(r)}
&=
\widetilde{\mathbf{Z}}^{(r)}
+
\operatorname{MHA}\!\left(
\operatorname{LN}_{Q}(\mathbf{Q}_A^{(r)}),
\operatorname{LN}_{E}(\widetilde{\mathbf{G}}^{(r)}),
\operatorname{LN}_{E}(\widetilde{\mathbf{G}}^{(r)})
\right),\\
\mathbf{Z}^{(r+1)}
&=
\mathbf{V}^{(r)}
+
\operatorname{FFN}\!\left(\operatorname{LN}_{V}(\mathbf{V}^{(r)})\right).
\end{aligned}
\label{eq:writeback}
\end{equation}
The residual is added to the anchor features \(\widetilde{\mathbf{Z}}^{(r)}\). The keys and values reuse \(\operatorname{LN}_E\) from the self-attention step.
Equations~\ref{eq:obs-read}--\ref{eq:writeback} together define the map \(\mathbf{Z}^{(r+1)}=\Phi_\theta(\mathbf{Z}^{(r)};\mathbf H,\mathbf{X}_A,\Lambda^{\mathrm{ref}},\mathbf c,\mathbf{X}_E^{(r)})\), applied for \(r=0,\ldots,R-1\) with shared parameters~\citep{UniversalTransformer}. Note that only \(\mathbf X_E^{(r)}\) and \(\mathbf{Z}^{(r)}\) change across steps. Figure~\ref{fig:evaluate-mix-write-back} illustrates stages (ii) and (iii).

\subsection{Output}
\label{sec:output}
For field prediction, the output decoder blends the final anchor features at each output query. It uses its own length scales \(\Lambda^{\mathrm{dec}}\) and corresponding weights \(w_j^{\mathrm{dec}}\) from Equation~\ref{eq:rbf}, giving \(\bar{\mathbf z}_{\mathrm{dec}}^{(R)}(\mathbf q)=\sum_jw_j^{\mathrm{dec}}(\mathbf q)\mathbf z_j^{(R)}\). A pointwise MLP maps this blend to an output value:
\begin{equation}
\widehat{g}(\mathbf{q})
=
D_{\theta}\!\left(\mathbf{q},\mathbf{X}_A,\mathbf{Z}^{(R)},\Lambda^{\mathrm{dec}},\mathbf c\right)
=
\operatorname{MLP}_\theta\!\left(\bar{\mathbf{z}}^{(R)}_{\mathrm{dec}}(\mathbf{q}),\gamma(\mathbf{q}),\mathbf{c}\right),
\label{eq:decoder}
\end{equation}
where \(\mathbf c\) is the optional global condition used during refinement. For vector outputs, we pool the final anchor features and apply an output head.

\subsection{Design Properties}
\label{sec:properties}
FST has the three properties below.

\noindent\textbf{Permutation invariance.}
Reordering encoded observation pairs leaves the prediction unchanged. Reordering evaluation points also leaves the anchor update unchanged. These follow from the set symmetries of attention~\citep{Perceiver}, allowing spatial samples to be processed independently of their storage order.

\noindent\textbf{Spatial localization.}
Spatial localization allows each output query to approximate the full prediction using only nearby anchors, whose features already contain global information from refinement. This reduces the number of features combined per query, with approximation error controlled by the omitted weight. Section~\ref{sec:local-decoding} evaluates this approximation on Burgers and Darcy.

\noindent\textbf{Query consistency.}
Once the final anchor features are computed, adding, removing, or reordering output queries leaves predictions at shared coordinates unchanged. Queries can therefore be decoded in batches or added in regions of interest while reusing the refined representation.

At fixed feature width and anchor count, self-attention scales quadratically with the number of evaluation points, while output decoding scales linearly with the number of queries. Appendix~\ref{app:computational-cost} gives the full computational cost.

\section{Experiments}
\label{sec:experiments}

We evaluate FST on Burgers and Darcy flow for field prediction and on ImageNet-1K for classification. Burgers ablations examine anchor coordinates and length scales, function-space updates, and recursive refinement. Each reported result comes from one final training run.

\subsection{Burgers Equation}
\label{sec:burgers-experiments}

\noindent\textbf{Setup.}
Each trajectory in the one-dimensional PDEBench Burgers dataset~\citep{PDEBench}
solves
\[
\partial_t u(t,x) + \partial_x \left(\frac{u(t,x)^2}{2}\right)
= \frac{\nu}{\pi}\partial_{xx}u(t,x),
\qquad x\in(0,1),\quad t\in(0,2],
\]
with periodic boundary conditions. Each solution is stored on a \(201\times1024\) time-space grid. The dataset covers 12 viscosities, with 24,000 training, 1,200 validation, and 1,000 test solutions. All models predict solution fields from the same initial conditions (ICs) and boundary conditions (BCs), which together comprise 1,424 values. Appendix~\ref{app:burgers-experiments} gives the experiment details. We evaluate predictions using the mean relative \(L_2\) error across solutions:
\begin{equation}
\label{eq:relative-l2}
\textstyle\overline{e}_{\mathrm{rel}}=\frac{1}{n}\sum_{i=1}^{n}\lVert\widehat{\mathbf{u}}_i-\mathbf{u}_i\rVert_2/\lVert\mathbf{u}_i\rVert_2,
\end{equation}
where \(\mathbf{u}_i\) and \(\widehat{\mathbf{u}}_i\) contain reference and predicted grid values for each of the \(n\) solutions.

We train on all 12 viscosities with viscosity as an additional input, and separately on \(\nu=0.01\) with 2,000 training, 100 validation, and 84 test solutions. Both FST models use four refinement steps.

\noindent\textbf{Comparisons.}
We compare FST with FNO~\citep{FNO} and Perceiver IO~\citep{PerceiverIO}. The single-viscosity experiment also includes Transolver~\citep{Transolver} and DeepONet~\citep{DeepONet}. FST and Perceiver IO use 2,500 supervised points per solution at each training step, while FNO uses the full grid. All use a batch size of 48. For each model, we select the checkpoint with the lowest validation error and evaluate it on the test set.

\begin{figure}[!t]
\centering
\includegraphics{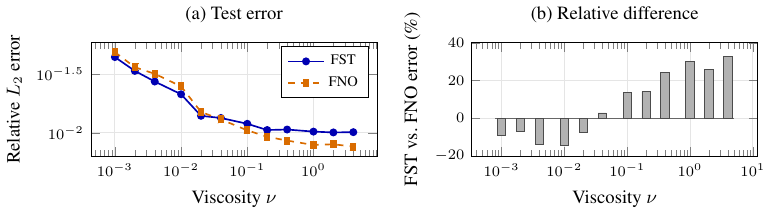}
\vspace{-0.1in}
\caption{Per-viscosity test error on PDEBench Burgers (all-viscosity models). (a) Relative \(L_2\) error. (b) \(100\,(e_{\mathrm{FST}}-e_{\mathrm{FNO}})/e_{\mathrm{FNO}}\); negative values favor FST. All 12 viscosities are shown.}
\label{fig:per-viscosity}
\end{figure}

\begin{figure}[!t]
\centering
\includegraphics[width=0.98\linewidth]{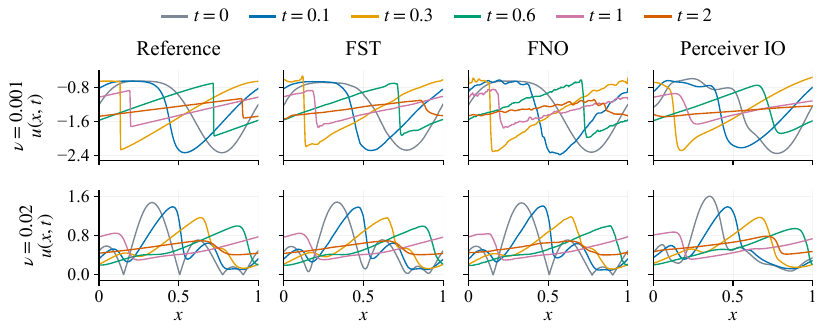}
\vspace{-0.1in}
\caption{Burgers profiles at \(\nu=0.001\) (top) and \(\nu=0.02\) (bottom). Colors indicate time; each row shares axis limits. FST captures fronts with fewer spurious oscillations (top) but rounds initial-state minima (bottom).}
\label{fig:burgers_profiles_main}
\end{figure}

\noindent\textbf{All-viscosity results.} Across 12 viscosities, FST substantially outperforms Perceiver IO and achieves error comparable to that of FNO (0.018180 versus 0.018528) at similar parameter counts (Table~\ref{tab:main-results}). Figure~\ref{fig:per-viscosity} shows that FST's error is 7.1--14.4\% lower than FNO's at the five lowest viscosities (\(0.001\)--\(0.02\); Table~\ref{tab:per-viscosity-results}). Lower viscosity produces sharper structures, which \citet{PDEBench} associate with increased difficulty for FNO. At \(\nu=0.001\), FST captures front locations and steepness with fewer spurious oscillations than FNO, while Perceiver IO broadens or displaces transitions (Figure~\ref{fig:burgers_profiles_main}). At \(\nu=0.02\), FST follows the later-time profiles but rounds sharp minima of the (observed) initial state, which FNO reconstructs more faithfully. Appendix~\ref{app:burgers_profiles} shows more cases.

\begin{table}[!t]
\caption{Burgers results across all 12 viscosities. Stop step: end of training. Selected steps (lowest validation error): FST 450k, FNO 400k, Perceiver IO 895k.}
\label{tab:main-results}
\centering
\small
\begin{tabular}{@{}lcrrr@{}}
\toprule
Model & Supervision per step & Parameters & Stop step & Relative \(L_2\) \\
\midrule
FST & 2,500 points & 2.901M & 500,000 & \textbf{0.018180} \\
FNO & full grid & 2.978M & 400,000 & \underline{0.018528} \\
Perceiver IO & 2,500 points & 2.995M & 900,000 & 0.073293 \\
\bottomrule
\end{tabular}
\end{table}

\noindent\textbf{Single-viscosity results.}
Across all four output grids, FST ranks second to FNO (Table~\ref{tab:single-viscosity-results}). FST's error stays near 0.042 across grids with fixed input observations. FST, Perceiver IO, and DeepONet evaluate the same predicted function at different points.

\begin{table}[!t]
\caption{Test relative \(L_2\) error on Burgers (\(\nu=0.01\)) at four output grids. See Appendix~\ref{app:burgers-experiments} for detailed inputs, outputs, and evaluation.}
\label{tab:single-viscosity-results}
\centering
\small
\setlength{\tabcolsep}{4pt}
\begin{tabular}{@{}lrrrrrr@{}}
\toprule
Model & Parameters & Supervision per step & \(32\times32\) & \(64\times64\) & \(128\times128\) & \(201\times1024\) \\
\midrule
FST & 2.831M& 2,500 points & \underline{0.04118} & \underline{0.04159} & \underline{0.04171} & \underline{0.04185}  \\
FNO & 2.978M & full grid & \textbf{0.02854} & \textbf{0.02950} & \textbf{0.02963} & \textbf{0.02974} \\
Perceiver IO & 2.937M & 2,500 points & 0.25634 & 0.25465 & 0.25396 & 0.25381 \\
Transolver & 2.956M & 2,500 points & 0.17709 & 0.07151 & 0.06569 & 0.06154 \\
DeepONet & 2.998M & 2,500 points & 0.24712 & 0.25204 & 0.25467 & 0.25581 \\
\bottomrule
\end{tabular}
\end{table}

\vspace{0.1in}
\subsection{Burgers Ablations}
\label{sec:ablation-studies}

All ablations use the single-viscosity Burgers task (\(\nu=0.01\)). Unless stated otherwise, they use \(32\times32\) anchors, adaptive coordinates, RBF length scales fixed at 0.03, four refinement steps, and the sampling settings in Appendix~\ref{app:provenance}. The \(32\times32\) model with adaptive coordinates and scales is the FST checkpoint in Table~\ref{tab:single-viscosity-results}. All the compared models are evaluated on the full grid.

\begin{table}[!t]
\centering\small
\caption{Ablation on anchor coordinates and length scales.}
\label{tab:anchor-ablation}
\begin{tabular}{@{}llrr@{}}
\toprule
Coordinates & Length scales & \(32^2\) anchors & \(16^2\) anchors \\
\midrule
Uniform & Fixed & 0.04604 & \underline{0.04571} \\
Adaptive & Fixed & \textbf{0.04069} & 0.05784 \\
Uniform & Adaptive & 0.05482 & 0.05094 \\
Adaptive & Adaptive & \underline{0.04185} & \textbf{0.04480} \\
\bottomrule
\end{tabular}
\end{table}

\begin{table}[!t]
\centering\small
\noindent\begin{minipage}[t]{0.43\linewidth}
\centering
\caption{Ablation on the evaluation, mixing, and write-back process.}
\label{tab:function-space-ablation}
\begin{tabular}{@{}lr@{}}
\toprule
Update & Error \\
\midrule
Evaluation, mixing, and write-back & \textbf{0.04069} \\
Two self-attentions on anchors & \underline{0.05212} \\
\bottomrule
\end{tabular}
\end{minipage}\hfill
\begin{minipage}[t]{0.54\linewidth}
\centering
\setlength{\tabcolsep}{4pt}
\caption{Ablation on recursive refinement.}
\label{tab:refinement-ablation}
\begin{tabular}{@{}ccr@{}}
\toprule
\shortstack{Refinement\\steps} & \shortstack{Attention to\\observations} & Error \\
\midrule
1 & Every step & 0.06209 \\
2 & Every step & \underline{0.04442} \\
4 & Every step & \textbf{0.04069} \\
8 & Every step & 0.06496 \\
4 & First step only & 0.05971 \\
\bottomrule
\end{tabular}
\end{minipage}
\end{table}
\noindent\textbf{Anchor coordinates and length scales.} The best configuration at each anchor count uses adaptive coordinates. At \(16\times16\), adapting coordinates and scales together gives the lowest error (0.04480), compared with 0.05784 for adaptive coordinates alone and 0.04571 for uniform coordinates and fixed scales. At \(32\times32\), learning with adaptive coordinates perform better.

\noindent\textbf{Evaluation, mixing, and write-back.}
We replace the evaluation, mixing, and write-back process with two self-attention blocks on the anchor features to test whether computing global interactions on the function domain gives better results. Table~\ref{tab:function-space-ablation} shows that this evaluation, mixing, and write-back process reduces error by 21.9\%, from 0.05212 to 0.04069. 

\noindent\textbf{Recursive refinement.}
Table~\ref{tab:refinement-ablation} shows that increasing the number of refinement steps can improve performance: four steps give the lowest error, 34.5\% below that of one step. However, increasing to eight steps harms performance. Attending to observations at every step reduces error by 31.9\% compared with attending only at the first step, showing its necessity in the refinement process.

\subsection{Darcy Flow}
\label{sec:darcy-experiments}

\noindent\textbf{Setup.}
We use 2,000 training, 100 validation, and 1,000 test solutions from the NeuralOperator Darcy dataset~\citep{NeuralOperatorDarcyDataset}. Models map coefficient fields to solution fields on a \(128\times128\) grid with full-grid supervision. We select checkpoints using a \(100\times100\) validation subset and report test relative \(L_2\) error on the full grid.

\noindent\textbf{Comparisons.}
We compare FST with FNO, Perceiver IO, Transolver, and DeepONet using the same tuning budget and convergence rule. Appendix~\ref{app:darcy-training} gives the model and training settings.

\begin{table}[!ht]
\centering\small
\begin{minipage}[t]{0.49\linewidth}\centering
\caption{Darcy flow: test relative \(L_2\) error on the full \(128\times128\) grid.}
\label{tab:darcy-results}
\setlength{\tabcolsep}{4pt}
\begin{tabular}{@{}lrrr@{}}
\toprule
Model & Params. & Sel.\ step & Rel.\ \(L_2\) \\
\midrule
FST & 2.831M & 58k & \underline{0.03173} \\
FNO & 2.911M & 42k & \textbf{0.03167} \\
Perceiver IO & 2.937M & 58k & 0.10102 \\
Transolver & 2.956M & 36k & 0.03671 \\
DeepONet & 2.993M & 18k & 0.14678 \\
\bottomrule
\end{tabular}
\end{minipage}\hfill
\begin{minipage}[t]{0.47\linewidth}\centering
\caption{ImageNet-1K validation accuracy at the selected checkpoints.}
\label{tab:imagenet-results}
\setlength{\tabcolsep}{4pt}
\begin{tabular}{@{}lrrr@{}}
\toprule
Model & Params. & Top-1 (\%) & Top-5 (\%) \\
\midrule
FST & 2,959,929 & \textbf{69.530} & \textbf{89.042}\\
ViT & 3,095,614 & 67.804 & 88.126\\
CNN-ViT & 3,158,150 & 61.052 & 83.478\\
\bottomrule
\end{tabular}
\end{minipage}
\end{table}

\noindent\textbf{Results.}
Table~\ref{tab:darcy-results} shows that FST achieves comparable error to FNO (0.03173 versus 0.03167) at a similar parameter count and outperforms the other baselines. Figure~\ref{fig:darcy-anchors-scales} in Appendix~\ref{app:darcy-training} shows the predicted anchor coordinates and decoding length scales.

\subsection{Decoding with Nearby Anchors}
\label{sec:local-decoding}
\begin{figure}[H]
\begin{minipage}[t]{0.50\linewidth}
\vspace{0pt}
Spatial localization allows each query to approximate the full prediction using fewer nearby anchors. We retain the \(k\) nearest anchors by Euclidean distance in the normalized domain during decoding and renormalize their RBF weights; refinement uses all 1,024 anchors. Figure~\ref{fig:local-decoding} shows the error approaching that of full decoding as the neighborhood grows. Burgers reaches nearly the same error with 16 neighbors, while Darcy requires a larger neighborhood. Evaluation uses 84 Burgers and 1,000 Darcy test cases. These results suggest that decoding from fewer anchors could reduce decoding cost.
\end{minipage}\hfill
\begin{minipage}[t]{0.47\linewidth}
\vspace{0pt}
\centering
\includegraphics[width=\linewidth]{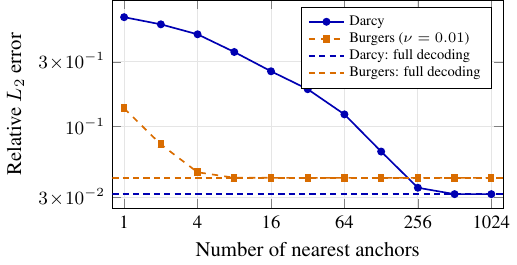}
\caption{Full-grid relative \(L_2\) error using different numbers of nearby anchors.}
\label{fig:local-decoding}
\end{minipage}
\end{figure}

\subsection{Image Classification}
\label{sec:imagenet-experiments}
Viewing images as color functions over space, we evaluate FST on image classification to test its effectiveness beyond physical fields.

\noindent\textbf{Setup.}
ImageNet-1K~\citep{ILSVRC15} contains 1,281,167 training images and 50,000 validation images. Each image produces 196 non-overlapping \(16\times16\) patches. For FST, each encoded patch is paired with its center coordinate. We report top-1 and top-5 accuracy.

\noindent\textbf{Comparisons.}
We train FST, ViT~\citep{ViT}, and CNN-ViT from scratch. FST and CNN-ViT use the same residual convolutional patch encoder; ViT uses a linear projection. All three train to convergence under the same stopping rule. Appendix~\ref{app:imagenet-training} gives their training settings.

\noindent\textbf{Results.}
FST reaches 69.530\% top-1 accuracy, 1.73 points above ViT, with fewer parameters (Table~\ref{tab:imagenet-results}). With the same patch encoder, CNN-ViT reaches 61.052\%. Appendix~\ref{app:imagenet-anchors} shows how the model distributes anchors across the image plane, with more anchors concentrating on relevant objects.

\section{Conclusion}
We proposed FST, which represents each input as a continuous latent field defined by adaptive anchors. Its function-space layer separates where features are stored from where global interactions are computed. Self-attention mixes features evaluated at sampled points, and cross-attention uses this information to refine the anchor features. Experiments on Burgers and Darcy flow demonstrate FST’s effectiveness for function prediction. Its performance on ImageNet-1K further demonstrates its ability to map input functions to finite-dimensional outputs. Ablations support the benefits of function-space updates and recursive refinement. The refined representation also supports queries at arbitrary coordinates, while spatial localization allows approximate decoding from fewer nearby anchors. Our experiments focus on rectangular domains. We expect to extend FST to irregular geometries by adapting anchor prediction and evaluation-point sampling to the domain.

\subsection*{AI use statement}
We used AI tools to polish the writing, help organize the paper, search for related work, and implement and run parts of the experimental code. We reviewed the AI-assisted work, checked the implementation, ran tests, and compared outputs with reference results. The authors take responsibility for the final paper, code, and reported results.

\begingroup
\small
\raggedright
\bibliography{references}
\bibliographystyle{abbrvnat}
\endgroup

\clearpage
\appendix
\raggedbottom
\section{Architecture Details}
\label{app:architecture-details}
\FloatBarrier
\subsection{Anchor Coordinates and Length Scales}
\label{app:anchor-representation}

\label{app:coordinate-parameterization}

We parameterize \(M=P(J+1)\) anchors on \(P\) fixed rows, with ordered, input-adaptive positions along each row. A transformer decoder predicts the spacing between anchors on each row. Its \(J\) learned query tokens attend to the encoded observations, each predicting one gap score for each of the \(P\) rows. We randomly initialize the tokens and learn them with the rest of the model.

\begin{figure}[htbp]
\centering
\includegraphics{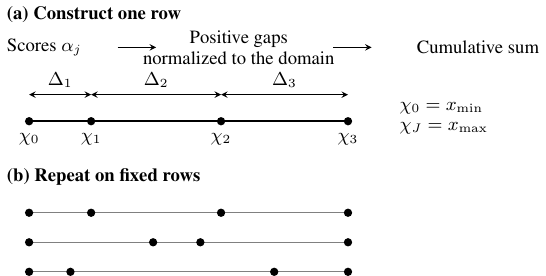}
\caption{Converting predicted gaps to anchor coordinates. Each row has its own predicted spacing.}
\label{fig:appendix-coordinates}
\end{figure}

For each row, we convert the scores \(\alpha_j\) into positive gaps and
normalize them so their sum equals the domain length:
\begin{equation}
\widetilde{\Delta}_j
=\frac{1}{2}\left(\alpha_j+\sqrt{\alpha_j^2+4}\right)+\varepsilon_g,
\qquad
\Delta_j=(x_{\max}-x_{\min})
\frac{\widetilde{\Delta}_j}{\sum_{m=1}^{J}\widetilde{\Delta}_m}.
\label{eq:appendix-gaps}
\end{equation}
Starting at \(x_{\min}\), we add the gaps successively to obtain the
\(J+1\) anchor coordinates:
\begin{equation}
\chi_0=x_{\min},
\qquad
\chi_j=x_{\min}+\sum_{m=1}^{j}\Delta_m,
\quad j=1,\ldots,J.
\label{eq:appendix-coordinates}
\end{equation}
The coordinates stay ordered, with the first and last anchors at the domain
endpoints. Figure~\ref{fig:appendix-coordinates} illustrates this construction.

\noindent\textbf{RBF length scales.}
For Burgers and Darcy, refinement and output decoding use separate learned base length scales for each anchor. Both use the same length-scale adjustments predicted from the input. For ImageNet, FST learns one scalar RBF length scale per anchor, shared across inputs. Coordinates and scales stay fixed across refinement.

\clearpage
\subsection{Model Configurations}
\label{app:task-configurations}

Tables~\ref{tab:fst-configurations} and~\ref{tab:fst-embeddings} give the FST dimensions and inputs to the coordinate predictor for each task. Sobol sequences are scrambled. The Burgers sampling counts use \(32^2\) anchors; Appendix~\ref{app:provenance} gives the \(16^2\) ablation.

\begin{table}[H]
\caption{FST architecture settings for the reported experiments.}
\label{tab:fst-configurations}
\label{tab:evaluation-sampling}
\centering\small
\begin{tabular}{@{}L{0.34\linewidth}L{0.19\linewidth}L{0.19\linewidth}L{0.19\linewidth}@{}}
\toprule
Setting & Burgers & Darcy & ImageNet-1K \\
\midrule
Anchors \(M\) & \(32\times32=1{,}024\) & \(32\times32=1{,}024\) & \(32\times32=1{,}024\) \\
Anchor feature dimension & 176 & 176 & 200 \\
Main attention heads & 8 & 8 & 8 \\
Observation-attention bias & Zero & Zero & Eq.~\ref{eq:imagenet-patch-read-bias} \\
Refinement steps \(R\) & 4 & 4 & 4 \\
Evaluation points \(S\) & 512 & 512 & 512 \\
Evaluation points: anchor / Sobol & 256 / 256 & 256 / 256 & 0 / 512 \\
Evaluation-point resampling & Every step & Every step & None \\
\midrule
Adaptive coordinate & Spatial & One spatial axis & Horizontal \\
Fixed rows \(P\) & 32 (time) & 32 & 32 (image) \\
Predictor query tokens \(J\) & 31 & 31 & 31 \\
Coordinate predictor width & 144 & 144 & 104 \\
Predictor attention heads / blocks & 4 / 2 & 4 / 2 & 4 / 2 \\
Positive-gap offset \(\varepsilon_g\) & 0.10 & 0.10 & 0.10 \\
Output-head weight / bias initialization & 0 / 0 & 0 / 0 & 0 / 0 \\
\midrule
RBF length-scale form & Axis-wise & Axis-wise & Scalar \\
RBF length-scale source & Input-predicted & Input-predicted & Learned per anchor \\
RBF length-scale initialization & 0.03 & 0.03 & 0.03 \\
RBF length-scale bounds & \([0.015,0.30]\) & \([0.015,0.30]\) & \([0.015,0.30]\) \\
Separate decoder length scales & Yes & Yes & n/a (pooled head) \\
\bottomrule
\end{tabular}
\\[2pt]
{\footnotesize The fixed-scale Burgers ablations use 0.03.}
\end{table}

\begin{table}[H]
\caption{Coordinate predictor encodings and observation inputs.}
\label{tab:fst-embeddings}
\centering\small
\begin{tabular}{@{}L{0.25\linewidth}L{0.23\linewidth}L{0.2\linewidth}L{0.23\linewidth}@{}}
\toprule
Component & Burgers & Darcy & ImageNet-1K \\
\midrule
Coordinate predictor & Fourier coordinate embeddings & Fourier coordinate embeddings & Fourier embeddings of patch-center coordinates \\
Predictor observation inputs & Coordinates, values, viscosity & Coordinates, coefficient values & Patch centers, patch features \\
\bottomrule
\end{tabular}
\end{table}

\FloatBarrier
\subsection{Computational Cost}
\label{app:computational-cost}

At fixed feature width, the leading-order cost of a dense implementation is
\[
O\!\left(C_{\mathrm{enc}}+C_{\mathrm{coord}}+R\,(MN+MS+S^2+M^2)+MQ\right).
\]
\(C_{\mathrm{enc}}\) and \(C_{\mathrm{coord}}\) denote the costs of the input encoder and coordinate predictor. The predictor's \(J\) query tokens attend to the \(N\) observations and predict \(PJ\) gap scores once per input, giving \(C_{\mathrm{coord}}=O(JN+J^2+PJ)\) at fixed width and depth.

Per refinement step, the attention to observations costs \(O(MN)\); evaluating the field at \(S\) points (Equation~\ref{eq:evaluation}) and the write-back cross-attention each cost \(O(MS)\); self-attention among evaluated features costs \(O(S^2)\); and forming the write-back queries \(\mathbf Q_A^{(r)}\) for all anchors costs \(O(M^2)\).

Decoding \(Q\) queries costs \(O(MQ)\) and reuses the final anchor representation, so for fixed \(M\) the decoding cost grows linearly in \(Q\).

\clearpage
\FloatBarrier
\section{Burgers Experiment Details}
\label{app:exp-details}
\label{app:burgers-experiments}

\FloatBarrier
\subsection{Data and Model Inputs}
\label{app:burgers-data}

The all-viscosity experiment uses all 12 viscosities; the single-viscosity experiment uses \(\nu=0.01\). All models receive the same ICs and BCs. The ICs contain the solution at \(t=0\) on all 1,024 spatial grid points. The BCs contain the solution values at the two spatial boundary grid points at each of the 200 later time steps. Together, they provide \(1{,}024+2\times200=1{,}424\) input values. Table~\ref{tab:burgers-data} summarizes the datasets and input sampling.

\begin{table}[H]
\caption{Burgers datasets and shared observation grid.}
\label{tab:burgers-data}
\centering\small
\begin{tabular}{@{}lrr@{}}
\toprule
Setting & All viscosities & \(\nu=0.01\) \\
\midrule
Training samples & 24,000 & 2,000 \\
Validation samples & 1,200 & 100 \\
Test samples & 1,000 & 84 \\
\midrule
Original grid (time \(\times\) space) & \multicolumn{2}{c}{\(201\times1024\)} \\
IC values & \multicolumn{2}{c}{1,024} \\
BC values & \multicolumn{2}{c}{\(2\times200=400\)} \\
\bottomrule
\end{tabular}
\end{table}

\subsection{Model Configurations}
\label{app:burgers-models}
\label{app:provenance}
\label{app:output}
\label{app:query-decoder}
\label{app:burgers-adapters}

\begin{figure}[htbp]
\centering
\includegraphics{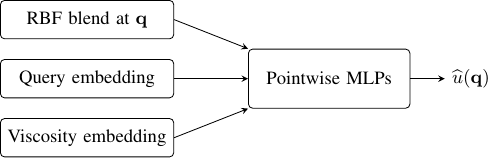}
\caption{Burgers decoding at one query coordinate.}
\label{fig:appendix-decoder}
\end{figure}

\noindent\textbf{FST output decoder.} For each query, we blend the final anchor features with RBF weights, then
use pointwise MLPs to predict the solution value. Query embeddings condition the prediction; the all-viscosity model also uses a viscosity embedding (Figure~\ref{fig:appendix-decoder}).
The output decoder has its own RBF length scales and MLP parameters.

\noindent\textbf{All-viscosity inputs.}
For the all-viscosity experiment, all models need to condition on the viscosity \(\nu\).
FST and Perceiver IO embed it as a global condition; FNO broadcasts it as a
constant input channel. In FST, a learned viscosity embedding is also added to the initial anchor features:
\[
\mathbf z_j^{(0)}=C_a(\gamma(\mathbf a_j))+C_c(\mathbf c).
\] Table~\ref{tab:model-inputs} compares the inputs
and supervision.

\begin{table}[H]
\caption{Model inputs and output supervision for the Burgers comparison.}
\label{tab:model-inputs}
\centering\small
\setlength{\tabcolsep}{4pt}
\begin{tabular}{@{}L{0.18\linewidth}L{0.42\linewidth}L{0.30\linewidth}@{}}
\toprule
Model & Input & Supervision per training step \\
\midrule
FST & ICs and BCs + \(\nu\) & 2,500 sampled points \\
Perceiver IO & ICs and BCs + \(\nu\) & 2,500 sampled points \\
FNO & Values + mask + \(\nu\) on the \(201\times1024\) grid & Full grid (205,824 points) \\
\bottomrule
\end{tabular}
\end{table}

\noindent\textbf{Baseline architectures.}
FNO and Perceiver IO use the same widths and depths in both Burgers experiments (Table~\ref{tab:burgers-baseline-config}). DeepONet uses branch widths \(1424\!\to\!918\!\to\!918\) and trunk widths \(2\!\to\!918\!\to\!918\), with ReLU activations and Glorot-normal initialization.

\begin{table}[H]
\caption{Burgers baseline architecture settings. Fourier modes follow the time-space axis order.}
\label{tab:burgers-baseline-config}
\centering\small
\begin{tabular}{@{}lrrrrr@{}}
\toprule
Model & Width & Layers & Heads & Latents/slices & Modes \\
\midrule
FNO & 48 & 4 & - & - & \(10\times30\) \\
Perceiver IO & 240 & 6 & 8 & 256 & - \\
Transolver & 328 & 5 & 8 & 32 & - \\
\bottomrule
\end{tabular}
\end{table}

\noindent\textbf{Single-viscosity inputs and outputs.} FNO and Transolver receive the same 1,424 IC and BC observations, without viscosity conditioning. We use the official FNO and irregular-mesh Transolver implementations with the following input and output adapters.

\noindent\textbf{FNO.}
We place the observed values on the original \(201\times1024\) time-space grid and fill unobserved entries with zeros. A second channel marks observed entries with one and unobserved entries with zero. FNO predicts the full grid and receives full-grid supervision. Its predictions at each evaluation grid are extracted from the original-grid prediction (Appendix~\ref{app:burgers-evaluation}).

\noindent\textbf{Transolver.}
We combine the observations \(\mathcal{O}=\{(\mathbf{x}_i,\mathbf{v}_i)\}_{i=1}^{N}\) and queries \(\mathcal{Q}=\{\mathbf{q}_k\}_{k=1}^{Q}\) into one point set. Each observation is represented as \((\mathbf{x}_i,\mathbf{v}_i,1)\), where \(\mathbf{v}_i\) is its standardized IC and BC value; each query is represented as \((\mathbf{q}_k,\mathbf{0},0)\). The final entry is a mask indicating whether the value is observed. Transolver predicts at all supplied coordinates; we retain and supervise only the query outputs.

\medskip
\noindent\textbf{FST configurations and ablations.}
Table~\ref{tab:provenance} gives the main FST configurations and the default ablation settings. Coordinate and length-scale ablations vary the coordinates and length scales as shown in Table~\ref{tab:anchor-ablation}. For both main Burgers models, the effective length scale of anchor \(j\) along axis \(d\) is
\begin{equation}
\ell^{s}_{j,d}=\operatorname{clip}\!\left(\exp\!\left(b^{s}_j+\delta_{j,d}(\mathcal O)\right),0.015,0.30\right),
\qquad s\in\{\mathrm{ref},\mathrm{dec}\}.
\label{eq:pde-length-scales}
\end{equation}
Here \(b^{s}_j\) is a learned log base scale, separate for refinement and decoding. The predictor supplies the shared axis-wise adjustment \(\delta_{j,d}=\log(4)\tanh(h_{j,d})\). Interior anchors average the outputs of the two adjacent gap tokens before this transformation; endpoints use the nearest token.
The two anchor counts use different numbers of sampled anchors: the \(32\times32\) model samples 256 anchors (25\%) and 256 Sobol points per step, whereas the \(16\times16\) model samples 128 anchors (50\%) and 384 Sobol points.

\begin{table}[H]
\caption{FST configurations for the Burgers comparisons and the default ablation. The default settings are varied in Tables~\ref{tab:anchor-ablation}-\ref{tab:refinement-ablation}.}
\label{tab:provenance}
\centering\small
\setlength{\tabcolsep}{5pt}
\begin{tabular}{@{}lclcr@{}}
\toprule
Experiment & Anchors & Coordinates & Length scales & \(R\) \\
\midrule
All viscosities & \(32^2\) & Adaptive & Adaptive & 4 \\
Single viscosity & \(32^2\) & Adaptive & Adaptive & 4 \\
Default ablation & \(32^2\) & Adaptive & 0.03 & 4 \\
\bottomrule
\end{tabular}
\end{table}

For the update ablation, two self-attention blocks on the anchor features replace stages (ii)-(iii) of the function-space layer. The replacement changes the features entering attention, the number of interaction tokens, and the write-back together, so the comparison supports the update block as a whole rather than isolating one of these factors.

\FloatBarrier
\subsection{Training and Checkpoint Selection}
\label{app:burgers-training}

All Burgers models minimize the mean squared per-solution relative \(L_2\) error in the original output units. Input values and prediction targets are standardized with shared scalar mean and standard deviation. The all-viscosity runs use \(-0.012048\) and \(0.621534\); the single-viscosity runs compute these statistics from their training targets on the \(100\times100\) grid.

Both Burgers FST comparisons sample 2,500 training queries per solution and training step from a \(100\times100\) subset of the original grid. We select the checkpoint with the lowest validation relative \(L_2\) error and evaluate it once on the test set.

\noindent\textbf{All-viscosity training.}
The all-viscosity comparison uses a shared convergence rule. After a minimum training
period, each model continues training until its best validation
relative \(L_2\) error improves by less than 1\% over 50,000 steps.
Table~\ref{tab:burgers-training} gives the training settings.

\begin{table}[H]
\caption{Shared training settings for the all-viscosity Burgers comparison.}
\label{tab:burgers-training}
\centering\small
\begin{tabular}{@{}ll@{}}
\toprule
Setting & Value \\
\midrule
Optimizer & Adam \\
Global batch size & 48 \\
Forward / backward precision & bfloat16 (BF16) \\
Loss accumulation precision & 32-bit floating point (FP32) \\
Minimum training steps & 250,000 \\
Maximum training steps & 1,000,000 \\
\bottomrule
\end{tabular}
\end{table}

All-viscosity FST uses learning rates \(10^{-3}\) for the main network and \(10^{-5}\) for the coordinate predictor. Both remain constant through step 200,000, decay by a cosine schedule to 5\% of their initial values at step 250,000, and remain constant thereafter. FNO uses the same schedule from \(5\times10^{-3}\); Perceiver IO uses a constant \(10^{-5}\).

All-viscosity FST selects step 450,000, with validation error 0.018014, and stops at step 500,000. FNO selects and stops at step 400,000; Perceiver IO selects step 895,000 and stops at 900,000.

\noindent\textbf{Single-viscosity training.}
FST and the four baselines use batch size 48, 1,000 warmup steps, validation every 2,000 steps, and at most 250,000 steps. The initial learning rate is \(10^{-3}\) for FST, \(3\times10^{-3}\) for FNO, and \(3\times10^{-4}\) for Perceiver IO, Transolver, and DeepONet. FST uses the same learning rate for its predictor and main network. If the best validation error improves by less than 1\% over 6,000 steps, the learning rate is divided by ten, up to two times. At the final learning rate, training stops when improvement is below 0.5\% over 12,000 steps, after at least 20,000 total steps.

\begin{table}[H]
\caption{Checkpoints for the single-viscosity models in Table~\ref{tab:single-viscosity-results}.}
\label{tab:burgers-baseline-checkpoints}
\centering\small
\begin{tabular}{@{}lrr@{}}
\toprule
Model & Selected step & Stop step \\
\midrule
FST & 92,000 & 98,000 \\
FNO & 72,000 & 72,000 \\
Perceiver IO & 34,000 & 50,000 \\
Transolver & 80,000 & 82,000 \\
DeepONet & 32,000 & 48,000 \\
\bottomrule
\end{tabular}
\end{table}

\noindent\textbf{Ablation checkpoints.}
Each ablation setting is trained separately. Table~\ref{tab:ablation-checkpoints} lists the selected steps for the settings in Table~\ref{tab:anchor-ablation}.

\begin{table}[H]
\caption{Selected training steps for the Burgers ablations; k denotes 1,000 steps. Left: coordinate and length-scale ablations at two anchor counts. Right: update and refinement ablations with \(32^2\) anchors.}
\label{tab:ablation-checkpoints}
\centering\small
\setlength{\tabcolsep}{4pt}
\begin{minipage}[t]{0.49\linewidth}\centering
\begin{tabular}[t]{@{}llrr@{}}
\toprule
Coordinates & Scales & \(32^2\) & \(16^2\) \\
\midrule
Uniform & Fixed & 82k & 116k \\
Adaptive & Fixed & 86k & 80k \\
Uniform & Adaptive & 130k & 112k \\
Adaptive & Adaptive & 92k & 78k \\
\bottomrule
\end{tabular}
\end{minipage}\hfill
\begin{minipage}[t]{0.48\linewidth}\centering
\begin{tabular}[t]{@{}lr@{}}
\toprule
Variant & Selected step \\
\midrule
Anchor self-attention & 108k \\
\(R=1\) & 76k \\
\(R=2\) & 74k \\
\(R=4\) & 86k \\
\(R=8\) & 74k \\
\midrule
\multicolumn{2}{@{}l}{Attention to observations} \\
First step only & 108k \\
\bottomrule
\end{tabular}
\end{minipage}
\end{table}

\noindent\textbf{Training cost.}
Table~\ref{tab:burgers-training-cost} reports the hardware, training time, and memory for the all-viscosity runs.

\begin{table}[H]
\caption{All-viscosity training on NVIDIA A100 80GB GPUs. Time sums the recorded training segments, including validation. Memory is the peak allocated memory per GPU.}
\label{tab:burgers-training-cost}
\centering\small
\begin{tabular}{@{}lrrr@{}}
\toprule
Model & GPUs & Time (h) & Memory (GiB) \\
\midrule
FST & 1 & 30.5 & 8.80 \\
FNO & 3 & 32.9 & 15.21 \\
Perceiver IO & 3 & 18.6 & 2.32 \\
\bottomrule
\end{tabular}
\end{table}

\FloatBarrier
\subsection{Evaluation}
\label{app:burgers-evaluation}

\noindent\textbf{Validation grids.}
For checkpoint selection in the single-viscosity comparison, FST and FNO use a \(100\times100\) subset of the original grid; Perceiver IO, Transolver, and DeepONet use \(64\times64\).

\noindent\textbf{Test grids and error.}
The all-viscosity FST checkpoint is evaluated in FP32 on the full \(201\times1024\) grid of all 1,000 test solutions. The single-viscosity comparison uses the four grids in Table~\ref{tab:single-viscosity-results}: \(32\times32\), \(64\times64\), \(128\times128\), and the original \(201\times1024\) grid. The ablations use the full original grid. For the single-viscosity comparison and ablations, Equation~\ref{eq:relative-l2} is computed per solution over the evaluation grid and averaged over the 84 test solutions.

Each output grid is a Cartesian product of original-grid subsets. For \(n\) points on an axis of size \(N\), we round \(\operatorname{linspace}(0,N-1,n)\) to integer indices and normalize them by \(N-1\).

\noindent\textbf{Model evaluation.}
FST, Perceiver IO, and DeepONet are queried directly at these coordinates. FNO predictions are extracted at the corresponding indices on the original grid. For Transolver, we permute the query indices with seed 2026091601 and partition them into groups of at most 2,500, each accompanied by all 1,424 observations. The four grids require 1, 2, 7, and 83 groups, respectively. Validation uses the same grouping procedure. All reported test evaluations use FP32. At each refinement step, FST draws 256 anchor indices without replacement and 256 scrambled Sobol points. Before each test-grid evaluation, we reset the anchor-selection seed to 2026091322 and the Sobol seed to 2026091323, keeping the sample order and batch size (8) fixed. This reproduces the same evaluation points across the four output grids. All query batches decode the resulting final anchor features.

\subsection{Additional Results}
\label{app:burgers-additional-results}
\label{app:per-viscosity}
\label{app:burgers_profiles}

\noindent\textbf{Per-viscosity results.}
Table~\ref{tab:per-viscosity-results} reports the relative \(L_2\) error within each viscosity group of the all-viscosity test set; Figure~\ref{fig:per-viscosity} plots the FST and FNO columns. FST has lower error than FNO at the five lowest viscosities, from \(0.001\) through \(0.02\); FNO has lower error at the seven higher viscosities. Perceiver IO has the highest error at every viscosity.

\begin{table}[H]
\caption{Per-viscosity relative \(L_2\) error on PDEBench Burgers. Best in bold; second best underlined.}
\label{tab:per-viscosity-results}
\centering\small
\begin{tabular}{@{}lrrr@{}}
\toprule
\(\nu\) & FST & FNO & Perceiver IO \\
\midrule
0.001 & \textbf{0.044195} & \underline{0.048623} & 0.147162 \\
0.002 & \textbf{0.033692} & \underline{0.036266} & 0.120887 \\
0.004 & \textbf{0.027352} & \underline{0.031828} & 0.112725 \\
0.01  & \textbf{0.021308} & \underline{0.024882} & 0.088623 \\
0.02  & \textbf{0.013957} & \underline{0.015056} & 0.063128 \\
0.04  & \underline{0.013382} & \textbf{0.013013} & 0.061419 \\
0.1   & \underline{0.011940} & \textbf{0.010499} & 0.059270 \\
0.2   & \underline{0.010567} & \textbf{0.009262} & 0.043572 \\
0.4   & \underline{0.010670} & \textbf{0.008588} & 0.044528 \\
1     & \underline{0.010242} & \textbf{0.007883} & 0.044719 \\
2     & \underline{0.010076} & \textbf{0.007991} & 0.041947 \\
4     & \underline{0.010131} & \textbf{0.007632} & 0.049411 \\
\bottomrule
\end{tabular}
\end{table}

\noindent\textbf{Solution profiles.} We extend Figure~\ref{fig:burgers_profiles_main} with profiles at additional viscosities and with different initial conditions. All examples come from the held-out test set and use the reported all-viscosity checkpoints. The four initial-condition examples are selected by profile shape before inference.

Models share the observations, six plotted times, and 1,024 spatial coordinates on the original grid. Curves show predictions, including at \(t=0\), in the original units. All three figures use the same column order and colors for each time. Within each row, the reference and all predictions share axis limits.

\begin{figure}[H]
\centering
\includegraphics[width=0.95\linewidth]{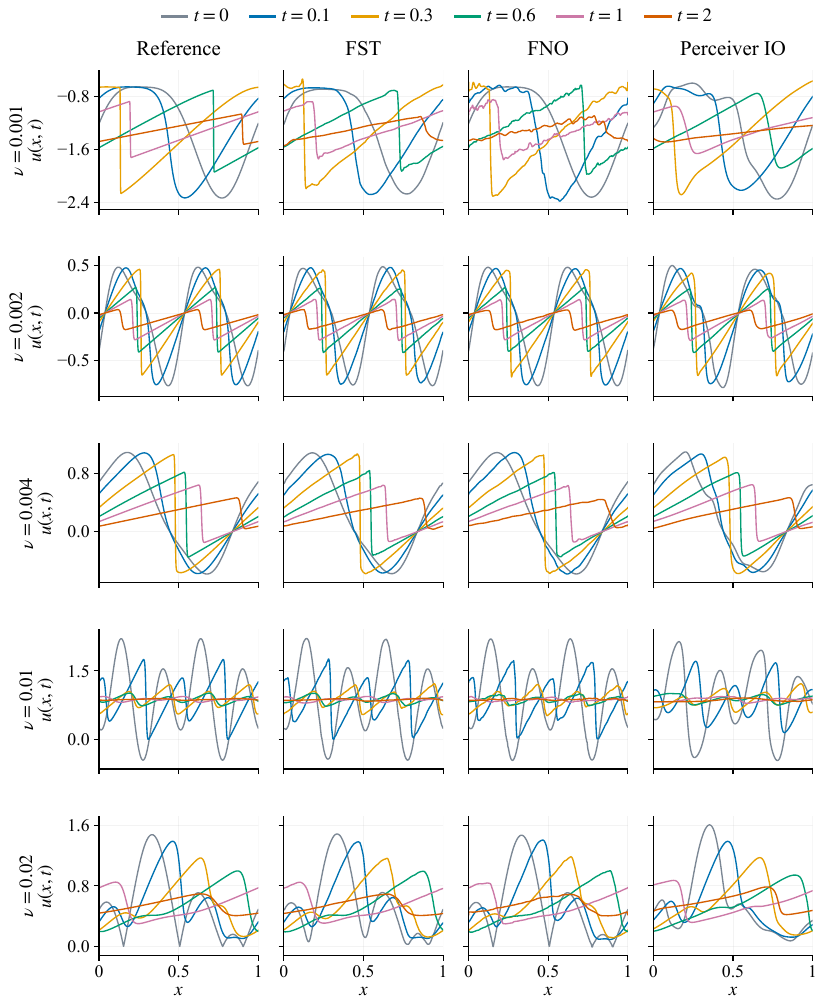}
\caption{Additional Burgers solution profiles. From top to bottom, rows show selected trajectories at \(\nu=0.001\), \(0.002\), \(0.004\), \(0.01\), and \(0.02\). Columns show the reference solution and predictions from FST, FNO, and Perceiver IO. Colors indicate \(t\in\{0,0.1,0.3,0.6,1,2\}\); axis limits are shared across columns within each row. The first and last rows are the examples in Figure~\ref{fig:burgers_profiles_main}. Different rows have different initial conditions.}
\label{fig:burgers_profiles_viscosities}
\end{figure}

\noindent\textbf{Profiles at selected viscosities.} Figure~\ref{fig:burgers_profiles_viscosities} compares selected trajectories at the five lowest viscosities. FST and FNO recover the main wave patterns across these examples, with local differences in front shape, spurious oscillations, and reconstruction of sharp initial-state features. Perceiver IO produces broader transitions and larger errors in oscillation amplitudes in several cases.

\begin{figure}[H]
\centering
\includegraphics[width=\linewidth]{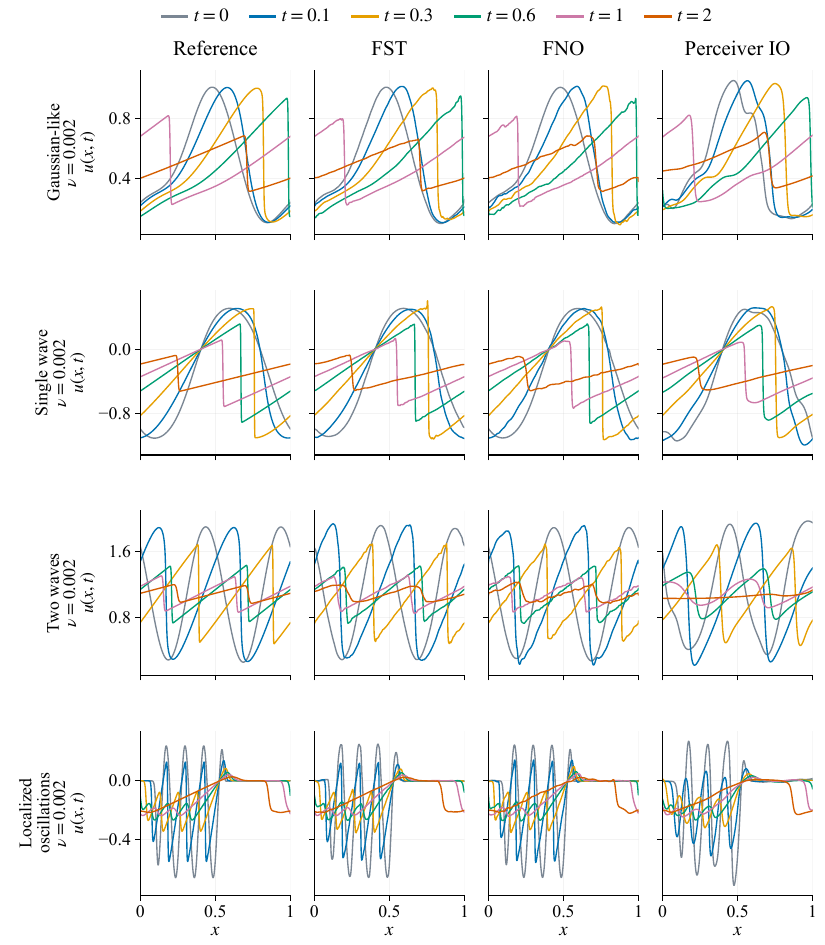}
\caption{Burgers solution profiles for different initial conditions at fixed viscosity \(\nu=0.002\). From top to bottom, the initial profiles are Gaussian-like, single wave, two waves, and localized oscillations. Columns show the reference solution and predictions from FST, FNO, and Perceiver IO. Colors indicate \(t\in\{0,0.1,0.3,0.6,1,2\}\); axis limits are shared across columns within each row. The examples contrast broad spatial features, repeated wave structures, and spatially concentrated oscillations.}
\label{fig:burgers_profiles_initial_conditions}
\end{figure}

\noindent\textbf{Different initial conditions.} Figure~\ref{fig:burgers_profiles_initial_conditions} compares four initial conditions at \(\nu=0.002\). FST and FNO recover the principal evolving fronts and oscillatory structures in these examples. FST exhibits fewer small fluctuations along several smooth segments, although local artifacts remain, including an overshoot near the \(t=0.3\) front in the single-wave example. Perceiver IO broadens several transitions and distorts some oscillation amplitudes. The example with localized oscillations shows that FST can preserve rapidly varying initial-state structure, even though it rounds some corner-like minima in the example of Figure~\ref{fig:burgers_profiles_main}.

\clearpage
\section{Darcy Experiment Details}
\label{app:darcy-training}

We use the NeuralOperator Darcy dataset~\citep{NeuralOperatorDarcyDataset}. Table~\ref{tab:darcy-setup} summarizes the data and training settings. We draw the training and validation splits from the 5,000-sample training file using seed 2026092001 and use all 1,000 samples in the test file. Input and output normalization statistics come from the training split.

\begin{table}[htbp]
\centering\small
\caption{Darcy data and training settings.}
\label{tab:darcy-setup}
\setlength{\tabcolsep}{6pt}
\renewcommand{\arraystretch}{1.12}
\begin{tabular}{@{}lrrl@{}}
\toprule
Split & Samples & Grid & Precision \\
\midrule
Training & 2,000 & \(128\times128\) & BF16 \\
Validation & 100 & \(100\times100\) & FP32 \\
Test & 1,000 & \(128\times128\) & FP32 \\
\bottomrule
\end{tabular}

\medskip
\begin{minipage}[t]{0.49\linewidth}
\centering
\begin{tabular}[t]{@{}lr@{}}
\toprule
Shared setting & Value \\
\midrule
Optimizer & Adam \\
Adam \(\epsilon\) & \(10^{-8}\) \\
Weight decay & 0 \\
Gradient clipping & 1 \\
Batch size & 48 \\
Warmup (steps) & 1,000 \\
Validation interval (steps) & 2,000 \\
\bottomrule
\end{tabular}
\end{minipage}\hfill
\begin{minipage}[t]{0.47\linewidth}
\centering
\begin{tabular}[t]{@{}lr@{}}
\toprule
Model & Selected learning rate \\
\midrule
FST & 0.001 \\
FNO & 0.005 \\
Perceiver IO & 0.0003 \\
Transolver & 0.001 \\
DeepONet & 0.001 \\
\bottomrule
\end{tabular}
\end{minipage}
\end{table}

FST uses the architecture settings in Table~\ref{tab:fst-configurations}. Its length scales follow Equation~\ref{eq:pde-length-scales}, with input-predicted adjustments along the two spatial axes.

Perceiver IO and Transolver use the widths, depths, and attention settings in Table~\ref{tab:burgers-baseline-config}. Darcy FNO uses four layers, width 50, and \(16\times16\) Fourier modes. DeepONet uses branch widths \(16384\!\to\!160\!\to\!256\!\to\!256\!\to\!256\) and trunk widths \(2\!\to\!256\!\to\!256\!\to\!256\!\to\!256\), with ReLU activations and Glorot-normal initialization.

FNO and Transolver process the coefficient and solution fields at the same grid locations. DeepONet takes the flattened coefficient field as its branch input and query coordinates as its trunk input.

The loss is the mean squared per-solution relative \(L_2\) error in original output units. FNO uses FP32 Fourier transforms during training.

For each model, we compare four learning rates over 4,000 steps, select the rate with the lowest mean of the last three validation errors, and train for an additional 8,000 steps to confirm the learning rate before training from scratch with seed 2026091311.

During final training, we smooth validation errors over three checks and halve the learning rate when the best smoothed error improves by less than 1\% over 4,000 steps, down to 1\% of the initial learning rate. At the minimum learning rate, training stops when improvement remains below 0.5\% over 12,000 steps.

Validation uses 100 approximately evenly spaced indices on the original grid along each axis. We select the checkpoint with the lowest unsmoothed validation error for testing. Test error is computed on the full \(128\times128\) grid. At each refinement step, FST draws 256 anchor indices without replacement and 256 scrambled Sobol points. At the start of each evaluation batch, it resets the anchor-selection seed to 2026391325 and the Sobol seed to 2026491325. The final anchor features serve all output queries.

\begin{figure}[htbp]
\centering
\includegraphics[width=\linewidth]{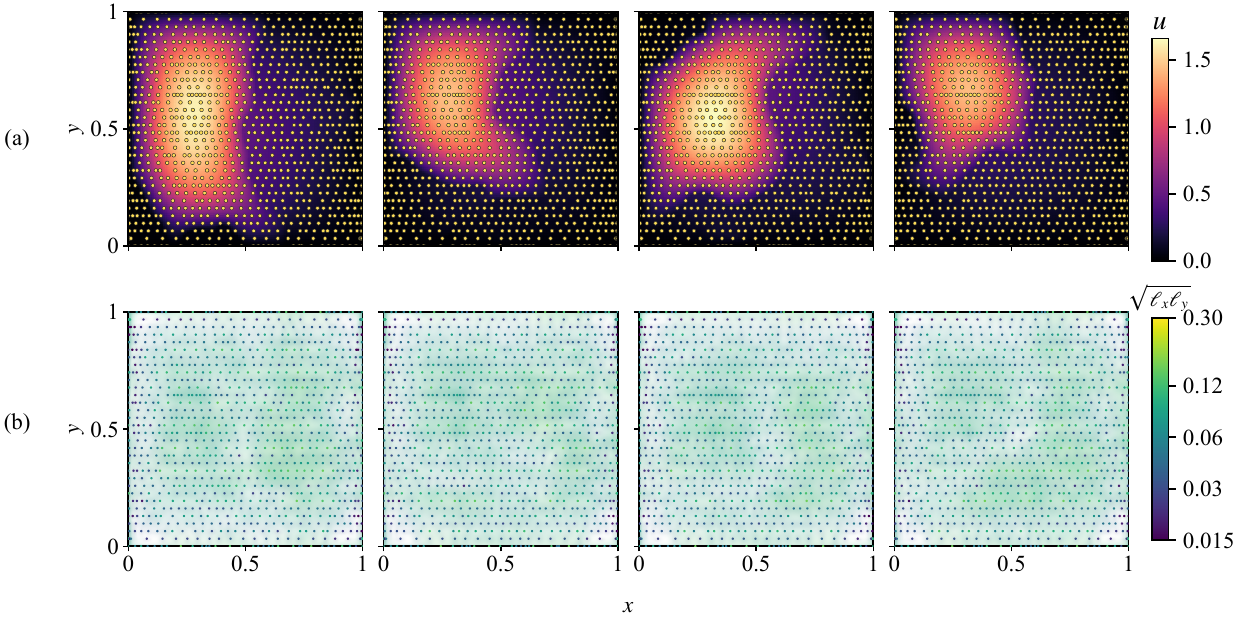}
\caption{Darcy anchor coordinates and decoding RBF length scales for four test samples. (a) Anchors over reference solutions. (b) Soft ellipses with horizontal and vertical scales proportional to \(\ell_x\) and \(\ell_y\), displayed at \(0.35\times\) scale; color indicates \(\sqrt{\ell_x\ell_y}\).}
\label{fig:darcy-anchors-scales}
\end{figure}

\FloatBarrier
\section{ImageNet-1K Experiment Details}
\label{app:imagenet-training}

FST and ViT train from scratch with the same data and optimization settings
for the comparison in Section~\ref{sec:imagenet-experiments}.

\FloatBarrier
\subsection{Dataset and Data Preprocessing}
\label{app:imagenet-data}

\begin{table}[H]
\caption{ImageNet-1K and data preprocessing.}
\label{tab:imagenet-data}
\centering\small
\begin{tabular}{@{}L{0.32\linewidth}L{0.63\linewidth}@{}}
\toprule
Setting & Value \\
\midrule
Training / validation images & 1,281,167 / 50,000 \\
Training crop & Random resized, \(224\times224\) \\
Training augmentations & Horizontal flips, 3-Augment~\citep{DeiTIII} \\
Color jitter & 0.3 \\
Random erasing probability & 0.25 \\
Validation resize (short side) & 256 \\
Validation center crop & \(224\times224\) \\
Validation augmentation & None \\
Non-overlapping patches & 196 of \(16\times16\) \\
\bottomrule
\end{tabular}
\end{table}

\subsection{Model Configurations}
\label{app:imagenet-models}
\label{app:imagenet-encoding}
\label{app:spatial-bias}

FST maps each patch to an observation token with a residual convolutional
encoder (Table~\ref{tab:patch-encoder}). The resulting tokens enter the
backbone specified in Table~\ref{tab:fst-configurations}.

\begin{table}[H]
\caption{FST patch encoder, in execution order. Spatial convolutions use
\(3\times3\) kernels and the encoder uses GELU activations.}
\label{tab:patch-encoder}
\centering\small
\begin{tabular}{@{}lll@{}}
\toprule
Stage & Operation & Setting \\
\midrule
1 & Convolution & 64 channels \\
2 & Residual blocks & 2 blocks, 2 convolutions each \\
3 & Pointwise projection & Width 200 \\
4 & Average pooling & Within each patch \\
5 & Add positional embedding & Fourier embeddings of patch bounds + MLP \\
6 & Residual feed-forward network & - \\
\bottomrule
\end{tabular}
\end{table}

\begin{figure}[H]
  \centering
  \includegraphics[width=\linewidth]{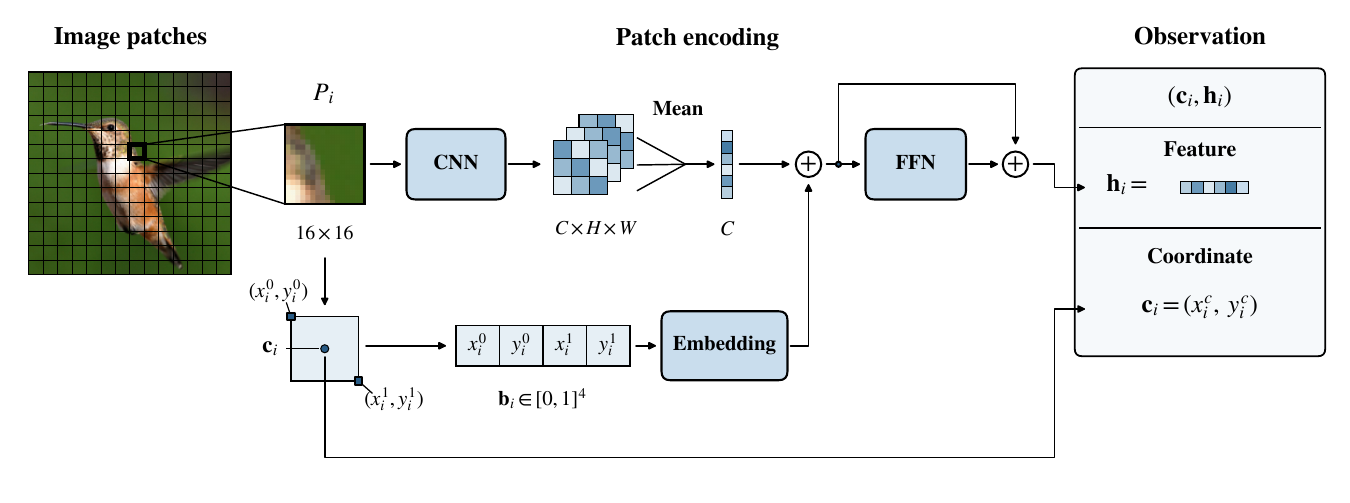}
  \caption{ImageNet patch observations. A CNN encodes each \(16\times16\)
  patch. Spatial pooling, a patch-bound embedding, and a residual FFN
  produce the feature \(\mathbf{h}_i\), paired with the patch center
  \(\mathbf{c}_i=((x_i^0+x_i^1)/2,(y_i^0+y_i^1)/2)\).}
  \label{fig:imagenet-patch-encoding}
\end{figure}

\noindent\textbf{Patch observations.} Figure~\ref{fig:imagenet-patch-encoding} shows the patch observation
\((\mathbf{c}_i,\mathbf{h}_i)\): a feature vector paired with its center
coordinate. Patch bounds are normalized to \([0,1]^2\).

\noindent\textbf{Attention bias.}
When anchors attend to patch features at the start of each refinement step, we add
\begin{equation}
B_{ji}=-\frac{\lVert\mathbf{a}_j-\mathbf{c}_i\rVert^2
+(w_i^2+h_i^2)/12}{2\sigma_{\mathrm{in}}^2}
\label{eq:imagenet-patch-read-bias}
\end{equation}
to the query-key logit in every attention head for anchor \(j\) and patch \(i\), where
\(w_i,h_i\) are the normalized patch width and height, and
\(\sigma_{\mathrm{in}}\) is a shared learned scalar, initialized to 0.20 and bounded in \([0.03,0.60]\). The numerator is the expected squared
distance from the anchor to a uniformly sampled point within the patch.
Because all patches have the same size, the size term adds the same constant to every logit and leaves the softmax weights unchanged. Attention weights depend jointly on this distance penalty and the learned
query-key similarity.

\noindent\textbf{Evaluation points.}
FST uses 512 fixed Sobol points in the image domain, independent of the patch grid.

\noindent\textbf{Classification head.} We apply layer normalization to each final anchor feature, average over
anchors, and use a linear classifier to predict the class logits.

\noindent\textbf{ViT baseline.} ViT uses a linear patch projection and the transformer configuration in
Table~\ref{tab:vit-configuration}.

\begin{table}[H]
\caption{ViT baseline configuration.}
\label{tab:vit-configuration}
\centering\small
\begin{tabular}{@{}ll@{}}
\toprule
Setting & Value \\
\midrule
Patch projection & Linear, \(16\times16\) patches \\
Feature width & 176 \\
Transformer layers & 10 \\
Attention heads & 4 \\
Feed-forward hidden width & 423 \\
Trainable parameters & 3,095,614 \\
\bottomrule
\end{tabular}
\end{table}

\noindent\textbf{CNN-ViT.} FST and CNN-ViT use the same residual convolutional patch encoder (stages 1-4 in Table~\ref{tab:patch-encoder}). CNN-ViT projects its output from width 200 to 176 and retains ViT's transformer blocks, positional embeddings, and classifier. It has 3,158,150 parameters.

\FloatBarrier
\subsection{Training and Evaluation}
\label{app:imagenet-optimization}
\label{app:imagenet-evaluation}
We train all models to convergence using tuned hyperparameters. Training stops when the best validation error improves by less than 1\% over the preceding 50,000 steps. The selected configuration and the learning-rate schedule until the convergence step are listed below.

FST and ViT use the optimizer settings in Table~\ref{tab:imagenet-optimizer}
and the staged learning-rate schedule in Table~\ref{tab:imagenet-schedule}.
FST's coordinate predictor uses one hundredth of the main learning rate
throughout training. CNN-ViT uses the same data preprocessing and optimizer settings. Its learning rate warms up to \(10^{-3}\) over 25,023 steps, then drops to \(5\times10^{-5}\) and follows a cosine decay toward \(2.5\times10^{-6}\) at step 500,000.

\begin{table}[H]
\caption{Shared ImageNet-1K optimization settings.}
\label{tab:imagenet-optimizer}
\centering\small
\begin{tabular}{@{}ll@{}}
\toprule
Setting & Value \\
\midrule
Optimizer & AdamW \\
Weight decay & 0.05 \\
Effective batch size & 256 \\
Computation precision & BF16 \\
Gradient norm clipping & 1.0 \\
Label smoothing & 0.1 \\
Base seed & 20260820 \\
\bottomrule
\end{tabular}
\end{table}

\begin{table}[H]
\caption{Learning-rate schedule for FST and ViT.}
\label{tab:imagenet-schedule}
\centering\small
\begin{tabular}{@{}llll@{}}
\toprule
Stage & Start & End & Learning rate \\
\midrule
Warmup & 0 & 25,023 & Peak \(10^{-3}\) \\
Cosine decay & 25,023 & 500,456 & \(10^{-3}\to 5\times10^{-5}\) \\
Cosine tail & 500,456 & 550,456 & \(5\times10^{-5}\to 0\) \\
Constant learning rate & 550,456 & 600,456 & \(2.5\times10^{-6}\) \\
\bottomrule
\end{tabular}
\end{table}

\clearpage
\subsection{Additional Anchor Layouts}
\label{app:imagenet-anchors}

\begin{figure}[H]
\centering
\includegraphics[width=\linewidth]{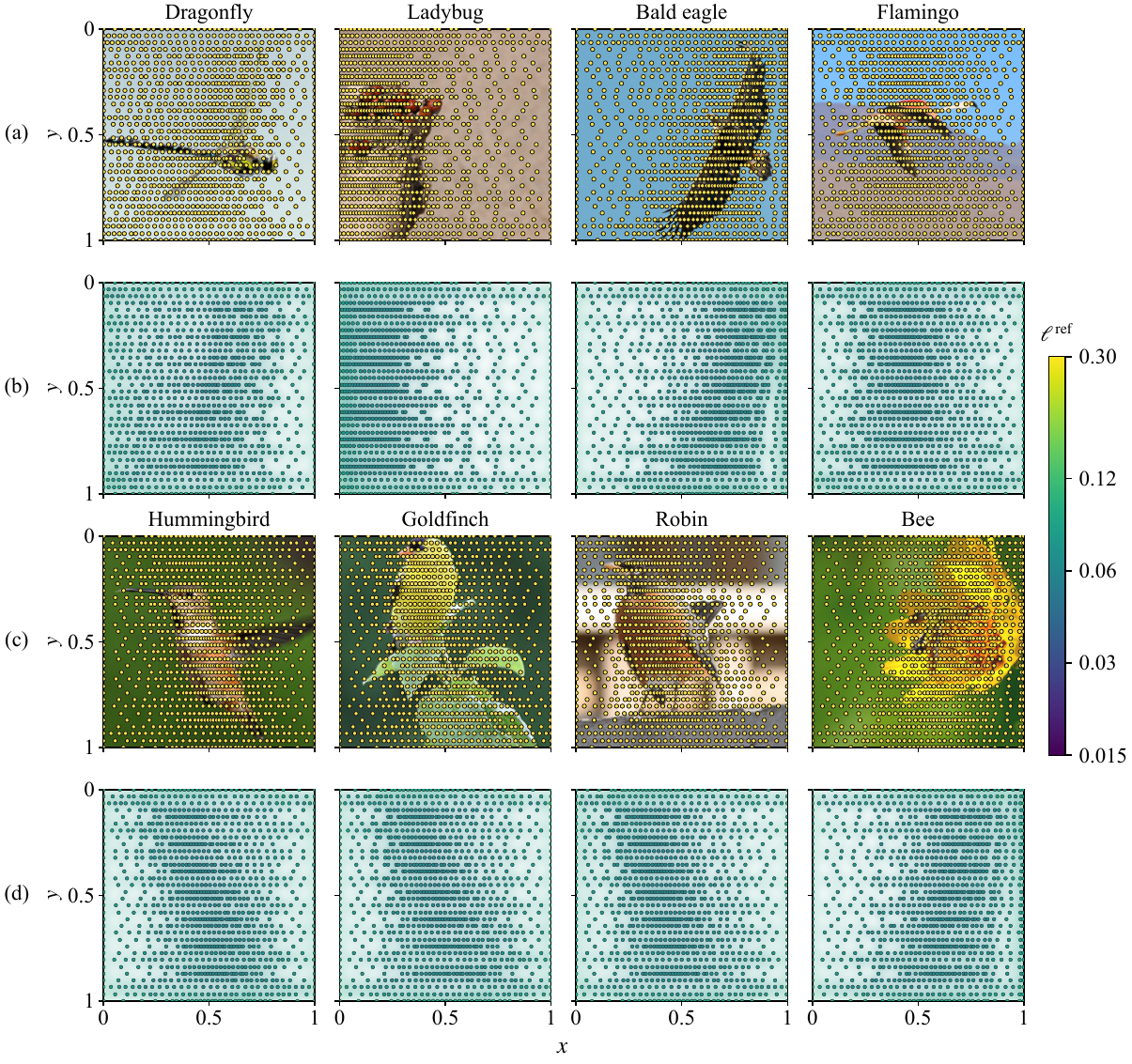}
\caption{Anchor coordinates and RBF length scales on ImageNet-1K validation images. (a, c) All 1,024 anchors overlaid on the images. (b, d) RBF length scales $\ell^{\mathrm{ref}}$, shown by color and soft circles displayed at $0.35\times$ scale. Horizontal coordinates adapt to each input; vertical rows are fixed.}
\label{fig:predicted_coordinates_on_imagenet}
\label{fig:appendix-imagenet-anchors}
\end{figure}

Figure~\ref{fig:predicted_coordinates_on_imagenet} shows anchor layouts and RBF length scales for eight validation images. The layouts vary across inputs.

\end{document}